\documentclass[runningheads]{llncs}
\usepackage{graphicx}
\graphicspath{ {./images/} }
\usepackage{tabulary}
\usepackage{latexsym}
\usepackage{float}
\usepackage{subcaption,booktabs}
\usepackage{microtype}
\usepackage{threeparttable} 
\usepackage{amsmath}
 \usepackage{enumerate}
\usepackage{amssymb}
\usepackage{pifont}
\usepackage{xcolor}
\usepackage{footnote}

\usepackage{longtable}

\usepackage[T5]{fontenc}
\usepackage[utf8]{inputenc}

\pdfoutput=1

\begin{document}

\title{Constructive and Toxic Speech Detection for Open-domain Social Media Comments in Vietnamese}

\author{Luan Thanh Nguyen\inst{1,2}\and
Kiet Van Nguyen\inst{1,2,}\thanks{Corresponding author}\and
Ngan Luu-Thuy Nguyen\inst{1,2}
\\\textit{17520721@gm.uit.edu.vn, kietnv@uit.edu.vn, ngannlt@uit.edu.vn}}

\authorrunning{Luan Thanh Nguyen et al.}
\titlerunning{Constructive and Toxic Speech Detection for Vietnamese Comments}
\institute{University of Information Technology, Ho Chi Minh City, Vietnam\and
Vietnam National University Ho Chi Minh City, Vietnam}

\maketitle 
\begin{abstract}
The rise of social media has led to the increasing of comments on online forums. However, there still exists invalid comments which are not informative for users. Moreover, those comments are also quite toxic and harmful to people. In this paper, we create a dataset for constructive and toxic speech detection, named UIT-ViCTSD (\textbf{Vi}etnamese \textbf{C}onstructive and \textbf{T}oxic \textbf{S}peech \textbf{D}etection dataset) with 10,000 human-annotated comments. For these tasks, we propose a system for constructive and toxic speech detection with the state-of-the-art transfer learning model in Vietnamese NLP as PhoBERT. With this system, we obtain F1-scores of 78.59\% and 59.40\% for classifying constructive and toxic comments, respectively. Besides, we implement various baseline models as traditional Machine Learning and Deep Neural Network-Based models to evaluate the dataset. With the results, we can solve several tasks on the online discussions and develop the framework for identifying constructiveness and toxicity of Vietnamese social media comments automatically.

\end{abstract}

\section{Introduction}
In the era of technology and the Internet, one of the most important factors that need to be concerned about is the quality of online discussions. Focusing on constructive comments and automatically classifying them promotes and contributes to improving online discussion quality and bringing knowledge for users. Besides, toxic comments, which cause the shutting down of user comments completely, are rising dramatically. Hence, filtering toxic comments relied on its level helps improve online conversations.

In this paper, we introduce the UIT-ViCTSD dataset for identifying constructiveness and toxicity of Vietnamese comments on social media. To ensure the quality of the dataset, we build a detailed and clear annotation scheme. Thereby, annotators can annotate comments correctly. Moreover, our system for detecting constructiveness and toxicity is proposed.

We organize the paper as follows. Section 2 reviews related works that are relevant to our works. Section 3 describes the dataset and defines the meanings of constructiveness and toxicity of comments. Section 4 introduces our proposed system for constructive and toxic speech detection. Section 5 describes experiments, results, and error analysis. Finally, Section 6 draws conclusions and future works.

\section{Related Work}
Constructiveness of comment is an essential element that positively gives users useful information and knowledge and promote online conversations. From these comments, conversations can be extended and more active, provide more information for users. Many Natural Language Processing (NLP) researchers all around the world has concentrated on characteristic. Since 2017, The New York Times has had full-time moderators to evaluate comments of users on their website. They then highlight comments that have constructiveness and label them as NYT Picks \protect\cite{nyt_picks} manually. In Japan, we can react like or unlike with comments of other users on Yahoo News. So that Yahoo filter their comments of users and rank that one by useful or not. A dataset named The Yahoo News annotated comments dataset was built to find good conversations online \protect\cite{ranking_dataset}. The keyword constructive was focused recently in 2017 by Napoles et al. \protect\cite{napoles_dataset}. They defined a new task for identifying good online conversations, called ERICs, which is used for classifying comments of users and identify good ones. 

In 2020, Varada et al. \cite{kolhatkar2020classifying} had an in-depth study about the constructiveness of comments, the main element promoting the quality of online conversations. They also built a dataset which is mainly about constructive comments named C3. Furthermore, they discussed the toxicity of comments and demonstrated the relationship between constructiveness and toxicity. They obtained the results of constructive comment detection with 72.59\% F1-score by the BiLSTM.

In recent years, NLP is increasing with the rise of high-quality datasets. For Vietnamese, there are several datasets about social media comments such as Vietnamese Social Media Emotion dataset (UIT-VSMEC) \cite{uit-vsmec} with 6,927 labeled comments and achieved a F1-score of 59.74\% by CNN model; Vietnamese Students’ Feedback dataset (UIT-VSFC) \cite{uit-vscf} with 16,000 labeled comments and the F1-score they gained by using Maximum Entropy was 84.03\%. Moreover, there are still datasets about preventing negative comments or hate speech, like the one presented by Vu et al. \cite{hatespeech}. However, Vietnamese is a low-resource language, and there is no dataset about the constructiveness of comments so far.  Hence, we aim to build a dataset to develop a framework for automatically identifying the constructiveness and toxicity of Vietnamese comments. With the results, we hope that we can use it to enhance the nature of online conversations, create meaningful and informative content, and make social media more useful.
\section{Dataset}
\subsection{Task Definition}
The main task in this paper is {\bf constructive speech detection} of Vietnamese comments. To understand the study, we explain constructive (as label 1) or non-constructive (as label 0) labels and their examples. Following the guidelines, annotators annotate comments with one of the two labels.

\begin{itemize}
    \item \textbf{Constructive:} Comments of users reinforce their point of view for the article.  Frequently, those comments provide lots of information and particularized opinions and contribute to promoting the topic.
    \item \textbf{Non-Constructive:} Comments of users which have little information. Their contents are only about expressing simple emotions and do not have much meaning.
\end{itemize}

Furthermore, we also concentrate on another task of \textbf{toxic speech detection}. With this characteristic, we annotate comments with one of four different labels consisting of very toxic, toxic, quite toxic, or non-toxic as follows:

\begin{itemize}
    \item \textbf{Very toxic:} Comments which have offensive contents; directly attacking individuals and organizations, showing disrespect to others. Especially using offensive words often causes shutting down conversation rapidly.
    \item \textbf{Toxic:} Comments whose contents are sarcasm and criticism; having a mockery attitude; disagreeing with the opinion but with a lack of delicacy and impolite.
    \item \textbf{Quite toxic:} Comments whose contents might be harmful to people (but not everyone) in specific contexts express disappointment.
    \item \textbf{Non-toxic:} Comments which have non-constructive content only express pure emotions and not make much sense.
\end{itemize}

Several examples of two tasks in our dataset are shown in Table \ref{tab:vidu_cons} and Table \ref{tab:vidu_tox} below.

\begin{table}[H]
\centering
\caption{Several examples about constructiveness of comments in our dataset. 1 = Constructive and 0 = Non-constructive.}
\label{tab:vidu_cons}
    \begin{tabular}{lp{10.5cm}c}
    \hline
    \textbf{No.} & \textbf{Comment} & \textbf{Label} 
            \\ \hline
    1   & Nội dung của bài viết là câu trả lời cuối đơn giản cho câu hỏi "Khủng hoảng tài chính là cái gì". Năm 2010,... Cuối cùng một câu trả lời đơn giản được đưa ra: "Bán mà không ai mua !"... cả thế giới đều thế thì khó mà mong đến việc giữ vững tốc độ tăng trưởng kinh tế! 
    
    \textit{(The content of this article is the right answer for "What is a financial crisis?". Finally, the most suitable answer is: "...")}
    & 1                           \\ 
    2   & Cầu mong ông sớm khỏe mạnh
    
    \textit{I hope you will get well soon}
    & 0                      \\ 
    3   & Tôi băn khoăn là ... mà không cần chữ kí của người mở thẻ hay sao ạ? Rồi còn phải .... Rồi còn phải làm sao nữa ạ? 
    
    \textit{I fret about ... without the owner's own signature? Then, ...? What should i do next? }
    & 1                      
            \\ \hline
    \end{tabular}
\end{table}

\begin{table}[H]
\centering
\caption{Several examples about toxic levels of comments in our dataset.}
\label{tab:vidu_tox}
    \begin{tabular}{lp{10cm}l}
    \hline
    \textbf{No.} & \textbf{Comment} & \textbf{Label}  
                \\ \hline
    1   & Đúng là cặn bã ấu dâm! 
    
    \textit{What a pedophile scum!}
    & Very toxic                     \\ 
    2   & Ích kỷ và vô ý thức. 
    
    \textit{Selfish and unconscious!}
    & Toxic                            \\ 
    3   & Xe Việt Nam vẫn quá đắt và đường xá thì quá kém
    
    \textit{Vietnamese cars are still too expensive and the quality of roads are too poor}
    & Quite toxic                      
                \\ 
    4   & 70\% cơ thể người là nước mà! 
    
    \textit{70 percent of the human body is made up of water} & Non-toxic               \\ \hline
    \end{tabular}
\end{table}

\subsection{Dataset Creation}
One of the most challenging issues we have to deal with is differences of Vietnamese language from other high-resource languages as English. A word may have different meanings in different sentences depending on the context. Also, its implications cause difficulties in building the Vietnamese dataset. Therefore, we have to refer to various other researches about the constructiveness of English comments and that definition in Vietnamese to build a high-quality dataset exactly and qualitatively.

\begin{figure}[H]
 \centering
 \includegraphics[width=1.\linewidth]{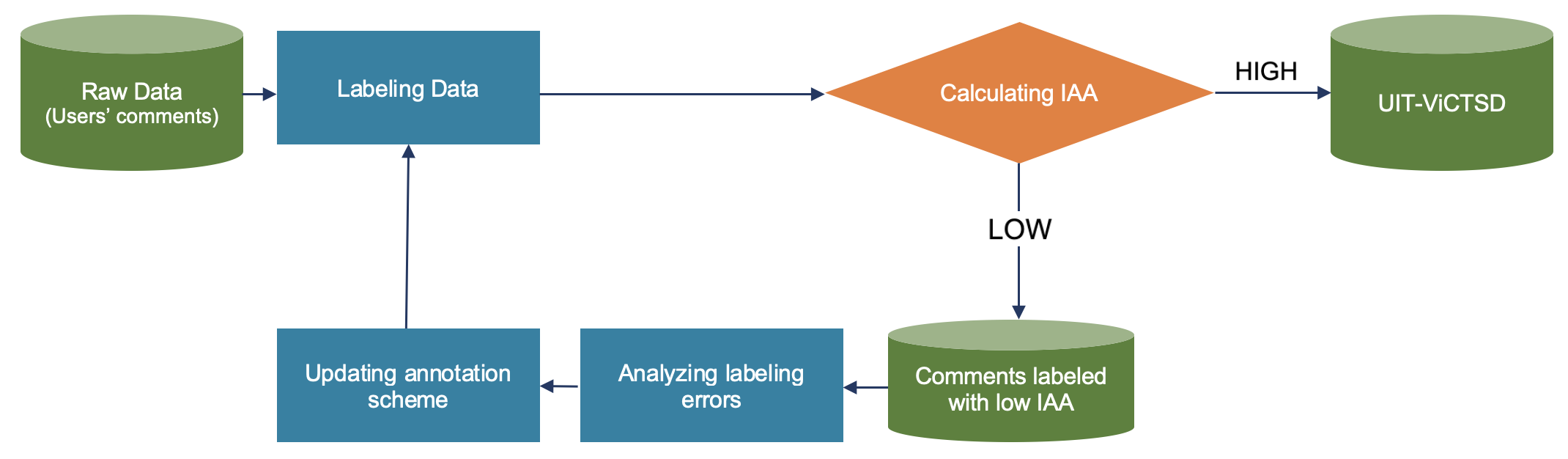}
 \caption{The process overview of building our annotated dataset.}
 \label{fig:progress}
\end{figure}

Figure \ref{fig:progress} shows us the progress of building this dataset with detailed steps. After crawling data, we start to label comments. The comment labeled with the low inter-annotator agreement (IAA) by members is labeled again as the given progression. Besides that, we analyze those cases and then update the annotation scheme again to improve the quality of the dataset.

Before collecting data, we do several surveys about our social media topic, which have many comments relevant to our interests. With the obtained results, we decide to crawl data from comments of users in VnExpress.net\footnote{https://vnexpress.net/} because there are huge comments that are useful for our research. Then, we build a tool for crawling data using Python and BeautifulSoup4\footnote{https://pypi.org/project/beautifulsoup4/} library useful functions for crawling data. The dataset consists of 10,000 comments and is divided equally into ten domains, including entertainment, education, science, business, cars, law, health, world, sports, and news (1,000 comments each domain)\footnote{http://nlp.uit.edu.vn/datasets/}. 

Three annotators label each comment in our dataset. Before labeling data, we built an annotation scheme with detailed and necessary information, which helps annotators label data quickly and precisely. We describe each task with a detailed definition. Several examples, as well as easily confused comments, are also listed in the annotation scheme. Therefore, annotators can label the data easier and more precisely. We also build a tool for labeling data easily and quickly.

After completing building the dataset, we split it into training, validation, and test sets with a 70:20:10 ratio by train\_test\_split function of scikit-learn\footnote{https://scikit-learn.org/}.

\subsection{Dataset Evaluation}
To evaluate the inter-annotator agreement of annotators, we use Fleiss' Kappa \protect\cite{agreementmeasure}. With that result, progress becomes more trustworthy.
\\

\begin{minipage}{.35\textwidth} %
    \begin{align*}
    A_m &= \frac{P_e - P_0}{1 - P_0} \tag{1} \label{eq:fleisskappa}
    \end{align*}
\end{minipage} %
\begin{minipage}{.6\textwidth} %
  Where:
\begin{itemize}
\item {$A_m:$ The inter-annotator agreement}
\item {$P_e:$ Expected probability of agreement among the annotators}
\item {$P_0$: Actual probability of agreement among the annotators}
\\
\end{itemize}
\end{minipage}
\\

To understand the definition of each label, annotators have to be trained strictly and carefully. Hence, we train them with five challenges (200 comments per round). After the challenges, members of the annotation team understand how to annotate comments with their labels. With the first challenge, the inter-annotator agreement is only 21.7\% in identifying constructiveness of comments and 30.4\% for toxic speech detection. The inter-annotation agreement in both tasks is not as high as expected in the first challenge because of differences in knowledge of each annotator in a separate domain. Thus, after each challenge, we consider conflicting cases, then we edit and update the annotation scheme. After five challenges, the inter-annotator agreement we gained for constructive and toxic speech detection is 59.48\% and 58.74\% IAA, respectively.

With the task of classifying constructive comments, the final label of a comment is chosen by the above 2/3 annotators. And the final label of comments in toxic speech detection is the average of the results of three annotators.

\subsection{Dataset Analysis}
\subsubsection{Distribution of Constructiveness}
\begin{figure}[!htb]
   \begin{minipage}{0.48\textwidth}
     \centering
     \includegraphics[width=1.0\linewidth]{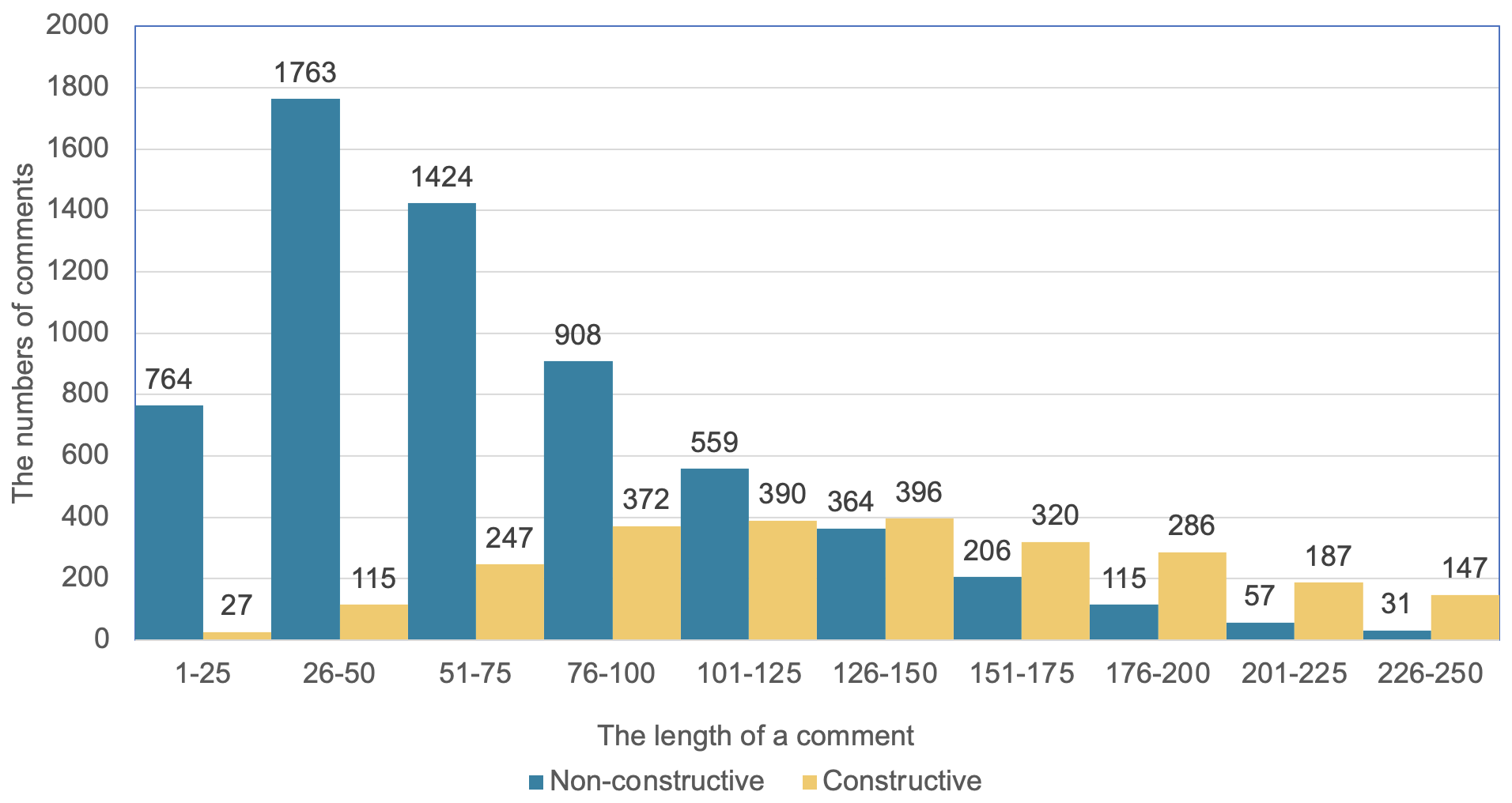}
     \caption{Distribution of constructive comments in the dataset.}\label{fig:sta_cons}
   \end{minipage}\hfill
   \begin{minipage}{0.48\textwidth}
     \centering
     \includegraphics[width=1.0\linewidth]{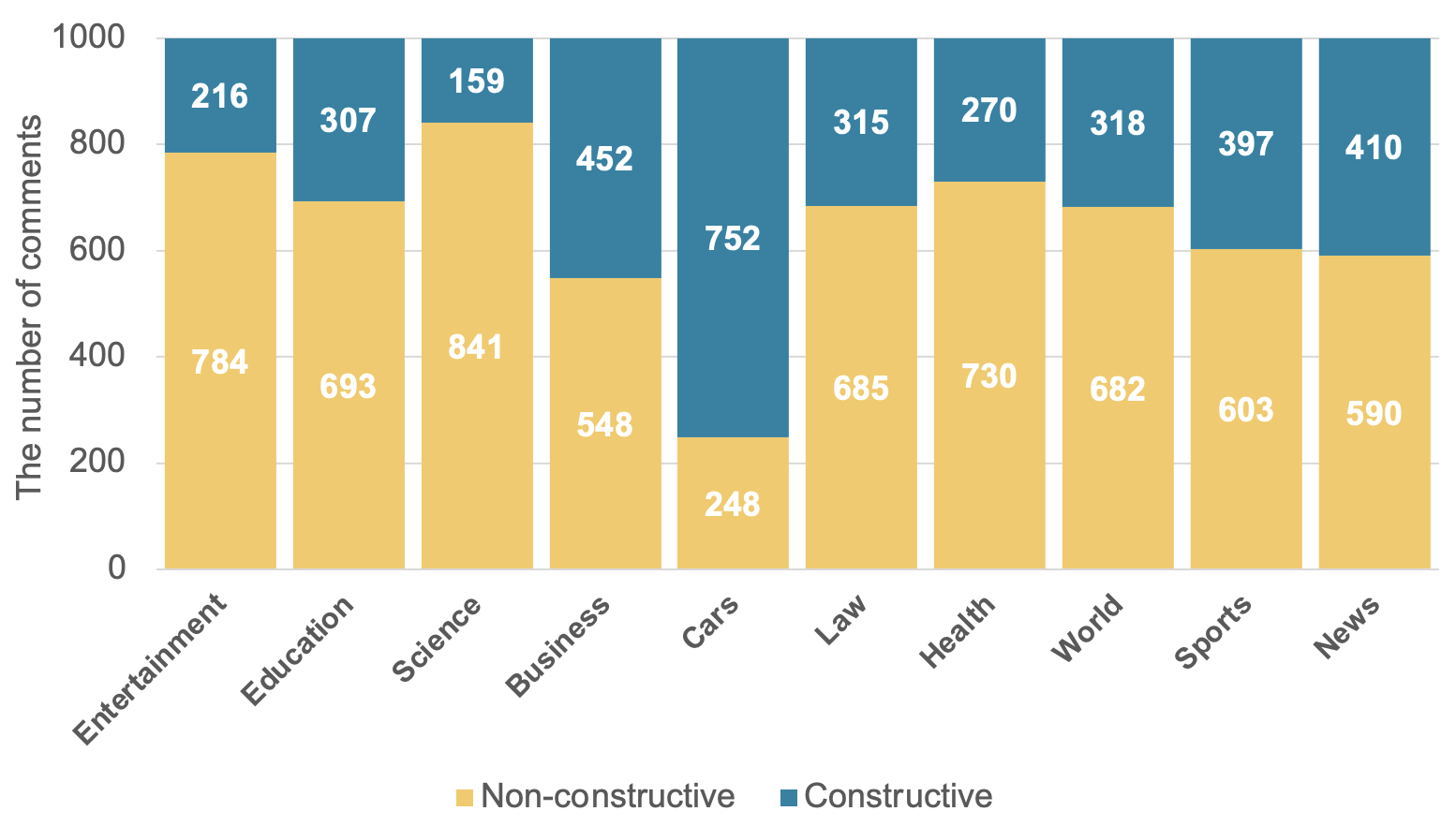}
     \caption{The number of constructive and non-constructive comments on each domain.}\label{fig:sta_cons_topic}
   \end{minipage}
\end{figure}

To gain an overview of the dataset, we conduct analyses on it. Figure \ref{fig:sta_cons} illustrates the distribution of constructive comments. We found that constructive comments are usually long comments and have a huge difference from non-constructive comments.


In addition, we analyze the number of constructive and non-constructive labels of comments on each domain to make an overview of the distribution of the dataset. The result is shown in Figure \ref{fig:sta_cons_topic}. Nevertheless, there are disadvantages we have to face. That is the imbalance of the dataset. The constructive comments are still less than non-constructive ones. Hence, we plan to balance the dataset in future.

\subsubsection{The relative between  Constructiveness and Toxicity}
\begin{figure}[H]
    \centering
    \includegraphics[width=.85\linewidth]{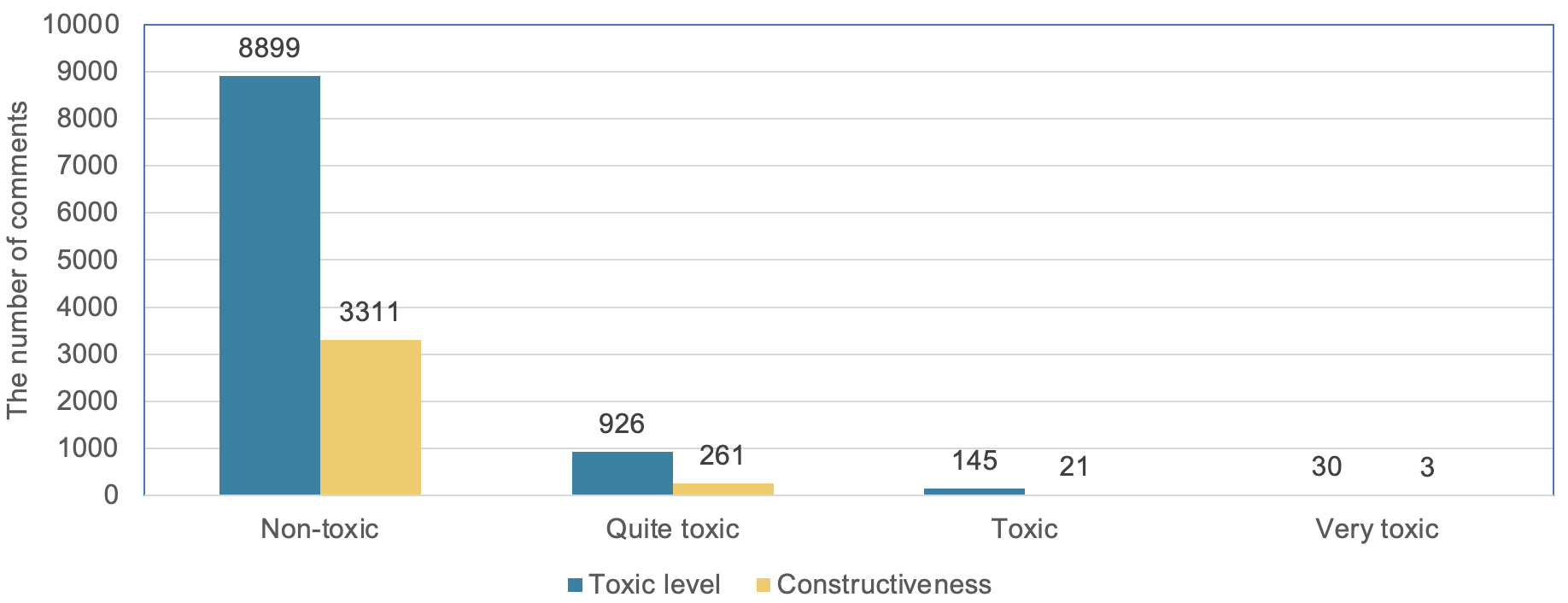}
    \caption{The number of constructive comments by their toxic levels.}
    \label{fig:cons_tox}
\end{figure}

We analyze the number of constructive comments by their toxic levels to obtain an overview of the relative between them, as shown in Figure \ref{fig:cons_tox}. We found that constructive comments also have toxicity. It proves that toxic contents still provide useful information for users.

Because the number of non-toxic comments is too large compared to the rest of the comments, we decide to combine the quite toxic, toxic, and very toxic labels into only one label as toxic. Hence, the task about toxic levels of comments becomes a problem with classifying a comment into toxic or non-toxic, a binary text classification task.
\section{Proposed System}
\subsection{System Overview}
\begin{figure}[H]
 \centering
 \includegraphics[width=1\linewidth]{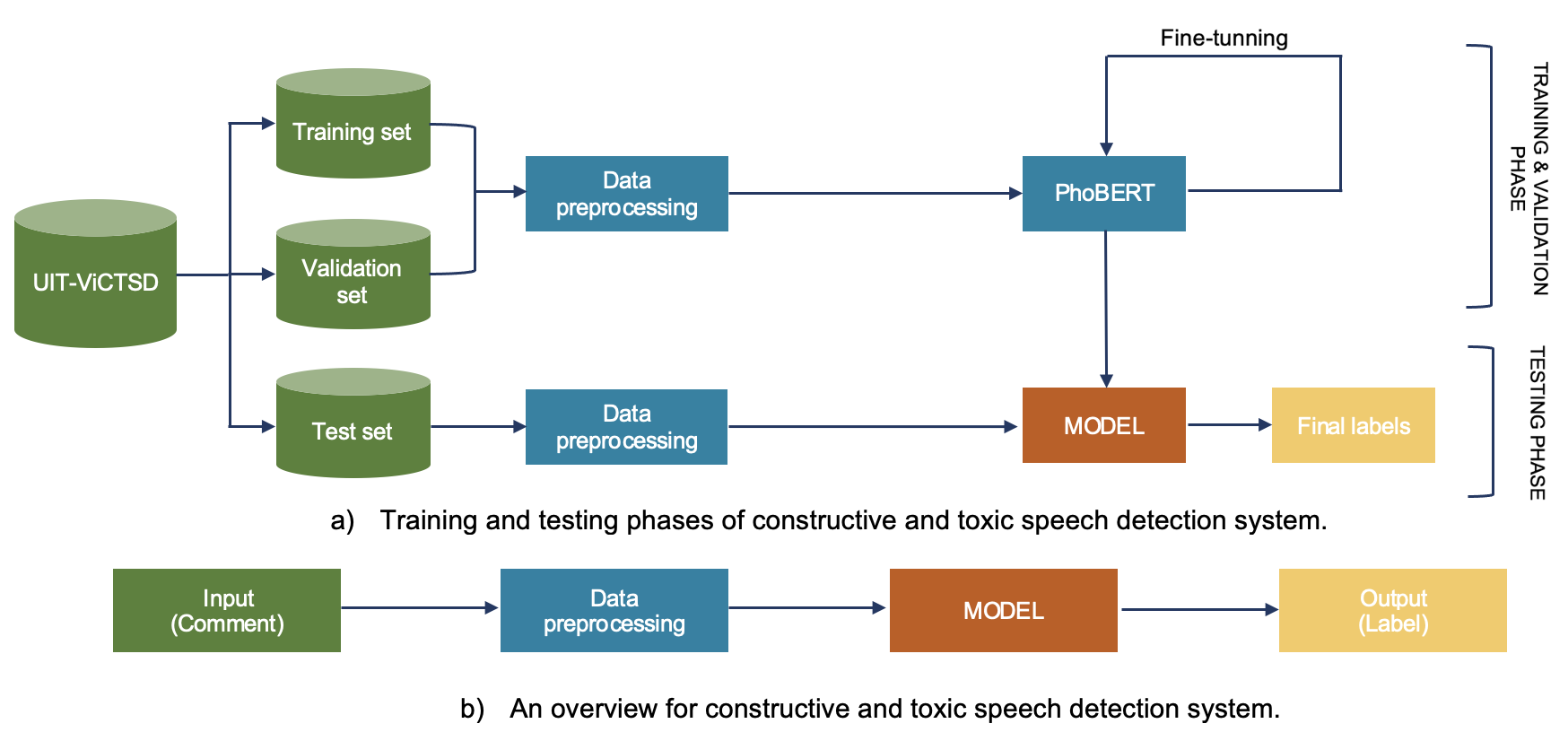}
 \caption{The proposed system for identifying constructiveness and toxicity of Vietnamse comments.}
 \label{fig:system}
\end{figure}

In this research, we propose a system for identifying the constructiveness and toxicity of Vietnamese comments. Figure \ref{fig:system} shows an overview of our proposed system described as follows.

\subsection{Data Preprocessing}
The dataset we built is a collection of texts, and it is an unstructured dataset, which is why it is difficult to be approached with models without applying preprocessing methods. Thus, we use popular techniques in NLP for preprocessing data of dataset. Firstly, we remove HTML tags. Unlike English or other popular languages, the space in Vietnamese is only the sign for separating syllables, not words. Tokenizing texts is important, and it directly affects the result of models. Hence, we tokenize those comments by using ViTokenizer of the library Pivi\footnote{https://pypi.org/project/pyvi/}.

\begin{longtable}[c]{ll}
\caption{Several examples about normalizing Vietnamese abbreviations.}
\label{tab:chuanhoatu}\\
\hline
\textbf{Abbreviations} & \textbf{Normalization} \\ \hline
\endfirsthead
\endhead
\hline
\endfoot
\endlastfoot
k, kh, hok, hk         & không                  \\
oke, okay, oki         & ok                     \\
vn, Vietnam, VN        & Việt Nam               \\ \hline
\end{longtable}

In the next step, we normalize texts by building a dictionary for abbreviations. Several examples are mentioned in Table
\ref{tab:chuanhoatu}.

Within the main model of our system, PhoBERT, which is mentioned in the next subsection, we use VNcoreNLP \protect\cite{vu2018vncorenlp} and FAIRSeq \protect\cite{ott2019fairseq} for preprocessing data and tokenizing words before applying to the dataset.

Finally, we eliminate spaces, especially special characters such as capital letters and symbols, before feeding them into training models. But, we still keep numbers because it also has essential information, which promotes constructive features.

\subsection{Transfer Learning - PhoBERT}
The transfer learning model has attracted increasing attention from NLP researchers around the world for its outstanding performances. One of the SOTA language models as BERT, which stands for Bidirectional Encoder representations from transformers, is published Devlin et al. \cite{devlin-etal-2019-bert}. It is a bi-directional transformer model for pre-training over lots of text data with no label to understand a language representation. Then, we fine-tune for specific problems.

For Vietnamese, the SOTA method was first released and called PhoBERT by Nguyen et al. \protect\cite{nguyen2020phobert} for solving Vietnamese NLP problems. PhoBERT is a pre-trained model, which has the same idea as RoBERTa, a replication study of BERT is released by Liu et al. \protect\cite{liu2019roberta}, and there are modifications to suit Vietnamese. In our proposed system, we use PhoBERT as the primary method for constructive and toxic speech detection for Vietnamese comments. We set the learning rate of 3e-5 and 24 for batch size.

\section{Experiments and Results}
\subsection{Baseline Systems}
Besides our system, we implement various models, including traditional methods as machine learning and neural network-based models.

\subsubsection{Machine learning models}

\begin{itemize}
    \item \textbf{Logistic Regression:} A basic machine learning algorithm for classification, especially binary task. 
    
    \item \textbf{Support Vector Machine (SVM):} This model is widely used for classification, regression and other problems in machine learning.
    
    \item \textbf{Random Forest:} An ensemble learning method for classification, regression, and other problems. It operates by constructing a multitude of decision trees at the training phase.
\end{itemize}

\subsubsection{Neural networks models}
\begin{itemize}
    \item \textbf{Long Short-Term Memory (LSTM):} This is a Deep Learning method based on RNN architecture. It is capable of learning long-term dependencies because of its feedback connections \protect\cite{gers1999learning}.
    
    \item \textbf{Bi-GRU-LSTM-CNN:} A customized ensemble model by Huynh et al. \protect\cite{van2019hate}. It is a combination of CNN-1D, Bidirectional LSTM and Bidirectional GRU layers and achieves optimistic results in binary classification task \protect\cite{van2020banana}.
\end{itemize}

Unlike machine learning methods, neural network models have different approaches to the dataset. Before the training model, we use the word-embedding for models as below.

\begin{itemize}
    \item \textbf{fastText:} A multilingual pre-trained word vector, including Vietnamese, released by Grave et al \protect\cite{grave-etal-2018-learning}.
    
    \item \textbf{PhoW2V:} PhoW2V is a pre-trained Word2Vec word embedding for Vietnamese, which is published by Nguyen et al. \protect\cite{phow2v_vitext2sql}. In this research, we use a PhoW2V word level embedding for preprocessing (300 dims).
\end{itemize}

\subsection{Experimental Settings}
For finding suitable parameters of each model, we use the GridSearchCV technique from sklearn. The final parameters we implemented are presented in Table \ref{tab:thamsomohinh}.

\begin{longtable}[c]{lp{4.25cm}p{4.25cm}}
\caption{Suitable parameters of each model for constructive and toxic speech detection of Vietnamese comments.}
\label{tab:thamsomohinh}\\
\hline
                    & \multicolumn{2}{c}{\textbf{Parameters}}                               \\ \hline
\endfirsthead
\endhead
\textbf{System}                    & \textbf{Constructiveness}         & \textbf{Toxicity}                 \\ \hline
Logistic Regression & C=100                             & C=100                             \\
SVM                 & C=1000, gamma=0.001,

kernel="rbf" & C=1000, gamma=0.001,

kernel="rbf" \\
Random Forest       & n\_estimators=108, max\_depth=400 & n\_estimators=108, max\_depth=400 \\ \hline
LSTM                &   units=128,

activation="sigmoid"                                & units=128,

activation="sigmoid"                                  \\
Bi-GRU-LSTM-CNN     &  CNN-1D: {drop\_out=0.2},

Bi-LSTM: {units=128, 

activation="sigmoid"}                               & CNN-1D: {drop\_out=0.2}, 

Bi-LSTM: {units=128, 

activation="sigmoid"}                                                        \\ \hline
\end{longtable}

\subsection{Experimental Results}
Before training models, we vectorize texts from the training set, from which models understand and learn from data. Then, we conduct experiments on our dataset, and the results of each model are shown in the Table \ref{tab:results} below. We use metrics as Accuracy and macro-averaged F1-score (\%) for evaluating the performances of models.

\begin{table}[]
\centering
\caption{The experimental results of each model for constructive and toxic speech detection.}
\label{tab:results}
\begin{tabular}{lcccc}
\hline
\multicolumn{1}{c}{\textbf{System}}                        & \multicolumn{2}{c}{\textbf{Constructiveness}} & \multicolumn{2}{c}{\textbf{Toxicity}} \\ \cline{2-5} 
                           & \textbf{Accuracy}     & \textbf{F1-score}     & \textbf{Accuracy} & \textbf{F1-score} \\ \hline
Logistic Regression        & 79.91                 & 70.78                 & 90.27             & 55.35             \\
Random Forest              & 79.10                 & 73.75                 & 90.03             & 55.30             \\
SVM                        & 78.00                 & 76.10                 & 90.17             & 59.06             \\ \hline
LSTM + fastText            & 80.00                 & 76.26                 & 88.90             & 49.63             \\
LSTM + PhoW2V              & 78.20                 & 77.42                 & 89.00             & 49.70             \\
Bi-GRU-LSTM-CNN + fastText & 79.90                 & 77.53                 & 89.10             & 48.88             \\
Bi-GRU-LSTM-CNN + PhoW2V    & 79.50                 & 77.94                 & 88.90             & 49.62             \\ \hline
\textbf{Our system}                 & 79.40                 & \textbf{78.59}        & 88.42             & \textbf{59.40}    \\ \hline
\end{tabular}
\end{table}

With the results above, our system achieves the best performance for identifying constructiveness of comments with the F1-score of 78.59\% and 59.40\% F1-score for toxic speech detection. However, the current results are not the highest ones. In particular, the F1-score result of the toxicity classification task is relatively low on the test set because of the significant imbalance between the toxic label and the non-toxic label, according to Figure \ref{fig:cons_tox}. Moreover, it proves that the dataset still has noisy data affecting classifier model performance directly.

\subsection{Error Analysis}
\begin{figure}[H]
 \centering
 \includegraphics[width=1\linewidth]{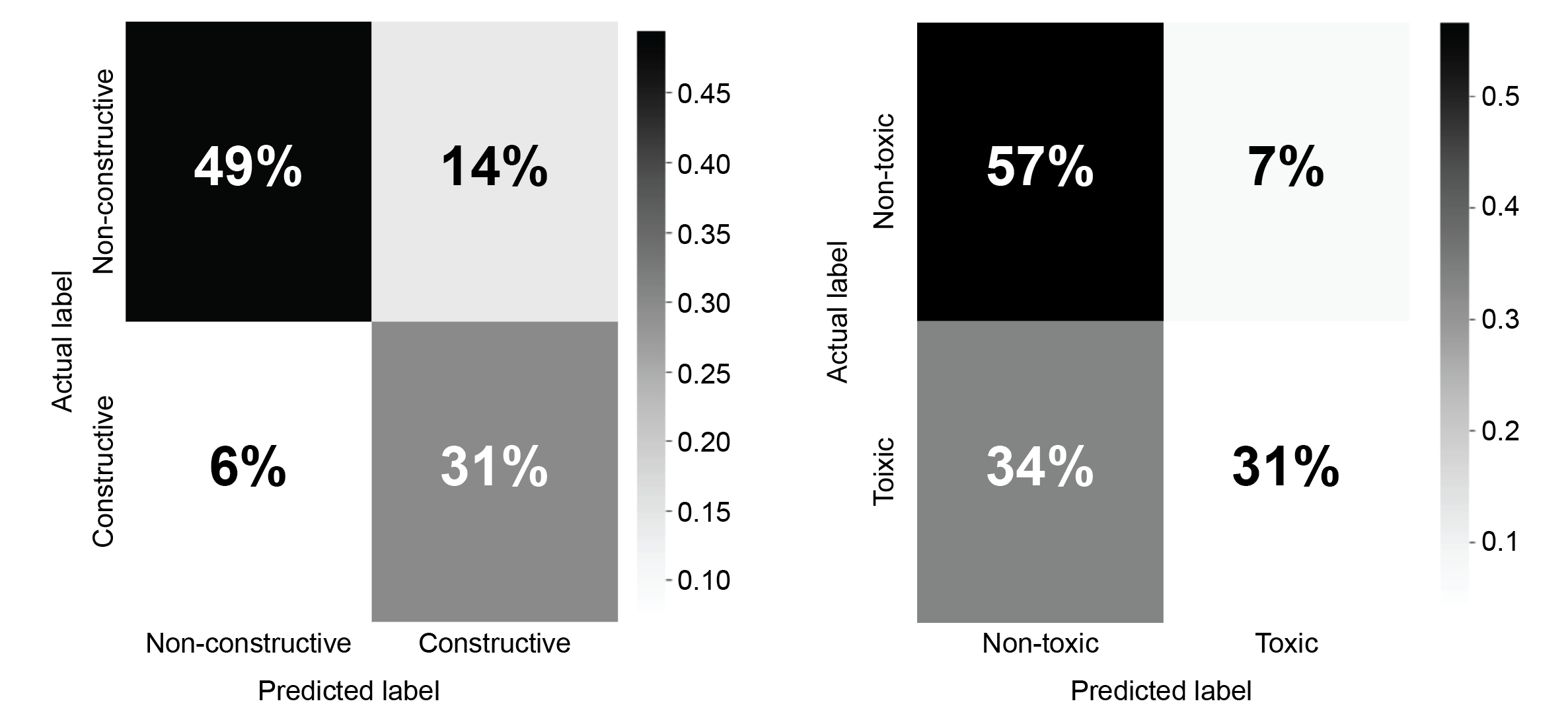}
 \caption{Confusion matrix of our system for constructive and toxic speech detection.}
 \label{Fig:cons_matrix}
\end{figure}

For analyzing errors of predictions by our system, we calculate a confusion matrix. Figure \ref{Fig:cons_matrix} presents the confusion matrix of our system for constructive and toxic speech detection. We see that in identifying constructiveness, the highest confusing cases are wrong to predict constructive (1) instead of non-constructive (0) with 14\% of the incorrect prediction. The confusion matrix of toxic speech detection shows us that the performance of the model for this task is still not high. The reason for that confusion is the imbalance of the dataset. Comments in our dataset are crawled mainly from official online sources, which blocks and prevents toxic contents on their forum. Hence, the number of toxic comments in the dataset are much less than the non-toxic ones, and models cannot be trained well and obtain the highest performances. We plan to improve its quality by adding more toxic comments from other social media platforms in the future.


\section{Conclusion and Future work}
This paper presented a new dataset named UIT-ViCTSD, including Vietnamese comments annotated manually with constructiveness and toxic labels. Firstly, we built an annotation scheme for labeling comments. In particular, we achieve the dataset consisting of 10,000 human-annotated comments on ten different domains. Secondly, we proposed a system for constructive and toxic speech detection with the state-of-the-art transfer learning model in Vietnamese NLP as PhoBERT. We achieved 78.59\% and 59.40\% F1-score for identifying constructive and toxic comments, respectively. Then, we implemented traditional machine learning and neural network models. Finally, we analyzed errors of our system on the dataset.

In future, we plan to evaluate the balanced dataset by removing inappropriately annotated comments and inserting new comments with the appropriate label. Besides, we intend to improve the quality of our dataset and our system in order to obtain the best results on constructive and toxic speech detection for Vietnamese social media comments.


\bibliographystyle{plain}
\bibliography{UIT-ViCTSD} 

\end{document}